\documentclass[sigconf,authorversion,nonacm,balance=false]{acmart}

\usepackage{fancyhdr}
\AtBeginDocument{%
	\addtolength{\footskip}{2.0\baselineskip}
	\fancyfoot[L]{\textbf{\textit{© Anna Volkova 2024. This is the author's version of the work. It is posted here for your personal use. Not for redistribution. The definitive version was published in International Conference on Information Technology for Social Good (GoodIT ’24), September 04–06, 2024, Bremen, Germany. https://doi.org/10.1145/3677525.3678651}}}
}

\usepackage{todonotes}
\usepackage[nolist]{acronym}
\usepackage{tabularx}
\AtBeginDocument{%
  }

\begin{document}

\title[Being Accountable is Smart]{Being Accountable is Smart: Navigating the Technical and Regulatory Landscape of AI-based Services for Power Grid}

\author{Anna Volkova}
\email{anna.volkova@uni-passau.de}
\orcid{0000-0001-9198-265X}
\affiliation{%
  \institution{University of Passau}
  \streetaddress{Innstraße 41}
  \city{Passau}
  \country{Germany}
  \postcode{94032}
}

\author{Mahdieh Hatamian}
\email{hatami02@ads.uni-passau.de}
\affiliation{%
  \institution{University of Passau}
  \streetaddress{Innstraße 41}
  \city{Passau}
  \country{Germany}
  \postcode{94032}
}

\author{Alina Anapyanova}
\email{alina.anapyanova@uni-passau.de}
\affiliation{%
  \institution{University of Passau}
  \streetaddress{Innstraße 41}
  \city{Passau}
  \country{Germany}
  \postcode{94032}
}

\author{Hermann de Meer}
\email{hermann.demeer@uni-passau.de}
\affiliation{%
  \institution{University of Passau}
  \streetaddress{Innstraße 41}
  \city{Passau}
  \country{Germany}
  \postcode{94032}
}

\renewcommand{\shortauthors}{Volkova et al.}

\begin{abstract}

The emergence of artificial intelligence and digitization of the power grid introduced numerous effective application scenarios for AI-based services for the smart grid. Nevertheless, adopting AI in critical infrastructures presents challenges due to unclear regulations and lacking risk quantification techniques. Regulated and accountable approaches for integrating AI-based services into the smart grid could accelerate the adoption of innovative methods in daily practices and address society's general safety concerns. This paper contributes to this objective by defining accountability and highlighting its importance for AI-based services in the energy sector. It underlines the current shortcomings of the AI Act and proposes an approach to address these issues in a potential delegated act. The proposed technical approach for developing and operating accountable AI-based smart grid services allows for assessing different service life cycle phases and identifying related accountability risks.

\end{abstract}

\ccsdesc[500]{Computing methodologies~Artificial intelligence}
\ccsdesc[500]{Security and privacy~Human and societal aspects of security and privacy}

\keywords{Accountability, Artificial Intelligence, Smart grid, AI regulations, Smart grid services}

\maketitle

\begin{acronym}
    \acro{SLA}{Service Level Agreement}
    \acro{AI}{Artificial Intelligence}
    \acro{GDPR}{General Data Protection Regulation}
\end{acronym}
\section{Introduction}
One of the main objectives of modern power grids is the uninterruptible supply of demands. At the same time, the power generation and distribution processes must be conducted safely to prevent any harm to society. The digitalization of the power grid has introduced significant risks, such as easier access to the power grid assets for cyber-attackers, but also substantial benefits, such as improved control systems and enhanced operational efficiency. System internal faults and external influences constantly challenge the power grid's resilience. The widespread adoption of \ac{AI} initiates the next phase of grid digitalization. Advanced smart grid services utilize \ac{AI} and underlying machine learning techniques to deliver insights beyond the reach of deterministic software and uncover correlations that may not be apparent to human operators. 

Some \ac{AI}-based services offer supplementary support, while others enhance or replace functions critical for maintaining the power grid's operation. Supplementary services include grid operation optimization, such as demand response, where \ac{AI} predicts fluctuations and reduces costs. As a part of the essential grid functions, \ac{AI} has also demonstrated being effective in forecasting renewable energy generation and load \cite{AHMAD2020loadforecasting}, stability analysis and control at different grid voltage levels \cite{SHI2020aistability}. Integrating such \ac{AI}-based services into the smart grid is associated with risks to the stable system operation and, as a result, to society. Data and algorithms used as the basis of \ac{AI}-based services might produce incorrect decisions and lead to operational failures, which can lead to blackouts. Excessive dependence on \ac{AI} can result in a lack of human oversight and hinder intervention in case of system failures. On the other hand, some \ac{AI}-based services, like failure detection, offer benefits by being capable of reacting much faster than human operators.

Safely implementing \ac{AI}-based services for the social good requires extensive regulation and methodologies to guarantee risk-free operation. The European Commission has attempted to address \ac{AI} integration through various regulatory initiatives and proposed \ac{AI} Act in 2021, which was adopted in March 2024 \cite{eu-AI} and complemented by a Corrigendum in April 2024 \cite{eu-Corrigendum}. Unfortunately, \ac{AI} Act does not provide domain-specific regulations, leaving space for interpretation regarding the \ac{AI} integration in the power grid. The central argument of this work is that regulating \ac{AI} integration in the smart grid requires a proper regulatory framework that can limit the risks related to \ac{AI}-based services operation and does not hinder the innovation in the smart grid domain. A technical framework should complement it to ensure precise risk quantification. Such a technical framework can be based on the concept of \ac{AI} accountability, defining it in a quantifiable manner, linking technical risks, their impact on the service operation, and responsible parties.

This work aims to support the development of the methodologies for regulated and accountable \ac{AI}-based smart grid services and makes recommendations for a potential delegated act for the smart grid. To achieve this, the following contributions are made: \textbf{1)} A classification of existing ways to address accountability and related terminology in the smart grid context, which enables the derivation of a quantifiable definition of \ac{AI}-based smart grid service accountability; \textbf{2)} An analysis of the current shortcomings of the \ac{AI} Act regarding the future risks of a narrow safety component definition for critical infrastructures; \textbf{3)} An approach for risk identification and accountability preservation as a part of a technical framework for developing and operating the accountable \ac{AI}-based smart grid service. The methodology focuses on assessing the different phases of \ac{AI}-based service development from the planning to model training and operation, identifying related accountability risks, and providing an overview of preservation measures. 

The paper is structured as follows. Section \ref{sec: related work} discusses existing research in accountable \ac{AI}-based services. Section \ref{sec: regulations} evaluates the existing regulations in the domain. Section \ref{sec: definitions} clarifies the concept of accountability, distinguishing it from related terms and providing a quantifiable definition. Section \ref{sec: framework} proposes a methodology to analyze accountability risks based on development methodologies.

  \begin{table*}[ht!]

\small{}
\begin{tabularx}{\textwidth}{p{0.5cm}>{\centering\arraybackslash}p{0.5cm}>
{\arraybackslash}p{7cm}>
{\centering\arraybackslash}p{1.8cm}>{\centering\arraybackslash}p{1.8cm}>{\centering\arraybackslash}p{1.8cm}>{\centering\arraybackslash}p{1.8cm}}
    
   \toprule

\textbf{Ref.} & \textbf{Year} & \textbf{Objective} & \textbf{Accountability} & \textbf{Power Grid} &\textbf{Technical}   & \textbf{Regulatory} \\

  \hline

\cite{floridi}& 2020 & Highlight critical ethical considerations for future AI & & & $\checkmark$ & \\

\cite{Krafft2020}& 2020 & Address risks and accountability issues for data subjects & $\checkmark$ & & & $\checkmark$ \\

\cite{ntoutsi} & 2020 & Provide a comprehensive overview of bias in AI systems &  & & $\checkmark$ & $\checkmark$ \\

\cite{kacian} & 2021 & Categorize and structure the definitions of accountability & $\checkmark$ &  &  & \\

\cite{Niet} & 2021 & Discuss EU AI Act's impact and risks on electricity sector & & $\checkmark$ & & $\checkmark$ \\

\cite{soares} & 2021 & Review authentication, authorization, and accountability& $\checkmark$ & $\checkmark$ & $\checkmark$ & \\

\cite{boch} & 2022 & An accountable AI framework with ethical and legal aspects & $\checkmark$ & & & $\checkmark$\\

\cite{Niet2022} & 2022 & Review regulations proposed by governmental institutions & & & & $\checkmark$ \\

\cite{paleyes2022challenges} & 2022 & Discuss the challenges in AI deployment phases & & & $\checkmark$ & \\

\cite{10.1145/3560107.3560253} & 2022 & EU AI Act vs. IEC 61508 for AI in critical infrastructure & & & $\checkmark$ & $\checkmark$ \\

\cite{volkova} & 2022 & Propose a monitoring system for AI-based smart grid & $\checkmark$ & $\checkmark$ & $\checkmark$ & \\

\cite{william} & 2022 & Make accountability and related concepts enforceable & $\checkmark$ & & & $\checkmark$ \\

\cite{diemert2023safety} & 2023 & Introduce safety integrity levels for AI in critical systems & & & $\checkmark$ &\\

\cite{hakem} & 2023 & Integrate Responsible Design Patterns into the AI pipeline& $\checkmark$ & & $\checkmark$ & \\

\cite{Hirvonen} & 2023 & Regulate accountability in automated decision-making& $\checkmark$ & & & $\checkmark$ \\

\bottomrule

 \end{tabularx}
 
    \caption{Analysis of related research from recent years based on the considered aspects}
    \label{tab:my_label1}
\end{table*}

\section{Related Work}

Some recent studies specifically address the definition of accountability and accountability requirements. For instance, in \cite{kacian}, the authors categorize and structure definitions of accountability, presenting a model to capture accountability in system design. In \cite{william}, authors discuss the need to narrow down high-level definitions of accountability and related terms and make these enforceable. In contrast, the present work focuses on sector-specific accountability and provides definitions that support quantification of accountability for \ac{AI}-based smart grid services. In \cite{boch}, accountability is discussed as the intersection of explainability and responsibility, and the existing regulations covering these concepts are analyzed. However, no approach to quantify accountability is proposed. While in \cite{boch} the authors only introduce the concept of implementable accountability, the present work proposes an approach to how implementable accountability can be achieved in the smart grid. 

Accountability in the smart grid domain has also been addressed from a technical point of view. Thus, in \cite{soares}, the authors review methods to provide authentication, authorization, and accountability in smart grids from the communication network perspective. In \cite{volkova}, the authors demonstrate a conceptual monitoring system for \ac{AI}-based smart grid services and indicate that monitoring of \ac{AI} in energy grid operations is essential for establishing accountability. In contrast, while showing the importance of monitoring, the present work proposes an approach to guarantee accountability for all the \ac{AI}-based service life cycle phases. 

The challenge of translating accountability into legal regulation has been discussed in \cite{boch}. In the absence of \ac{AI}-specific regulations, IEC 61508 \cite{iec61508-1} has been reviewed as a source of recommendation for \ac{AI} integration in critical infrastructures. Thus, authors in \cite{10.1145/3560107.3560253} state that all \ac{AI} applications in the critical infrastructure are generally not recommended in 
the standard. However, the Standard only discourages using \ac{AI} for particular applications in safety-related functions. An attempt to adjust Satefy Integrity Levels for \ac{AI} has been discussed in \cite{diemert2023safety}. The proposed approach is valid but too abstract: \ac{AI} safety risk classification should be done more granularly.   

One of the core insights of the present work is that accountability is only achievable through accountable development. The impact of the development flaws on the \ac{AI}-based service performance has been discussed in \cite{ntoutsi} with a focus on data bias. The present work considers bias mitigation, among other steps, as an important accountability guarantee. An extensive survey on potential design flaws in \ac{AI} development pipeline is presented in \cite{paleyes2022challenges}. The present work refines it for smart grid applications and discusses the role of risk identification as an accountability mechanism. In \cite{hakem}, a survey on responsible \ac{AI} development process and required tools are presented. While the collected responses can support the design of the toolbox for risk reduction at different phases, the accountability of each tool should be analyzed separately. 

Further related work is summarized in Table \ref{tab:my_label1}. Most studies focus on accountability in \ac{AI} or specifically within power systems. Additionally, some authors discuss the regulatory aspects of \ac{AI} and analyze recent regulations proposed by the EU and various governmental institutions alongside the technical assessment of accountability in \ac{AI}. In \cite{Niet}, additional use cases of \ac{AI} in the electricity sector, such as forecasting models and flexibility asset management, are discussed but were not yet identified under a particular risk category of the \ac{AI} Act or analyzed in the context of the safety component definition. This paper follows up on this discussion with a more detailed analysis.

\label{sec: related work}

\section{AI-based services in the smart grid}

\label{sec: regulations}

A smart grid has been defined by many different institutions and authors, for example, as ``\textit{an advanced digital two-way power flow power system capable of self-healing, adaptive, resilient, and sustainable with foresight for prediction under different uncertainties}'' \cite{DILEEP20202589smartgridsurvey}. Many of these advanced features elevate the system from its pure physical purpose of transferring the power flow to an intelligent system capable of reacting and overcoming the existing challenges, improving its cost-effectiveness and customer satisfaction. \ac{AI} has been demonstrated effective in essential technical, including safety-related functions, and user- and grid economy-oriented services. Some functions, such as voltage and frequency control, directly impact the system operation. Demand response, while being a critical function for the system's operation effectiveness, is not strictly required for system operation. Client-oriented functions, such as consumption analysis and billing optimization, do not affect the system's operation directly. The services also function on different time scales, allowing or hindering human oversight. Control and fault mitigation services make decisions quickly during the operation, while forecasting services provide calculations in advance and ensure sufficient time for operators to react to the proposed predictions. In this work, the term \ac{AI}-based smart grid service considers both technical and process optimization functions operating within the smart grid and having \ac{AI}-based components. This work does not discuss information security-related services.

To guarantee the safe operation of the \ac{AI}, the responsive regulations should consider the level of \ac{AI} involvement in the operation of the particular smart grid function. This section discusses the existing regulations in the \ac{AI} domain and their applicability to the power grid domain. After the in-depth analysis of the most important formulations, the problem of deriving regulations and recommendations and the need for \ac{AI} accountability is discussed.

\subsection{Current Status of AI Regulation in Energy}

\ac{AI} and Machine Learning technologies have increasingly come to the attention of European lawmakers in recent years. The result was a consensus at the European level that a common regulatory framework was needed to both foster \ac{AI}-related innovation and to contain its risks in several fields. The European \ac{AI} Act, adopted in March 2024 and being part of the European digital strategy, constitutes, therefore, a novel European legislative instrument aiming to ensure a \textit{safe, transparent, traceable, non-discriminatory, and environment-friendly application of \ac{AI}} technology \cite{eu-AI}. 

\begin{figure*}[t!]
    \centering
    \includegraphics[
  width=0.6\linewidth
]{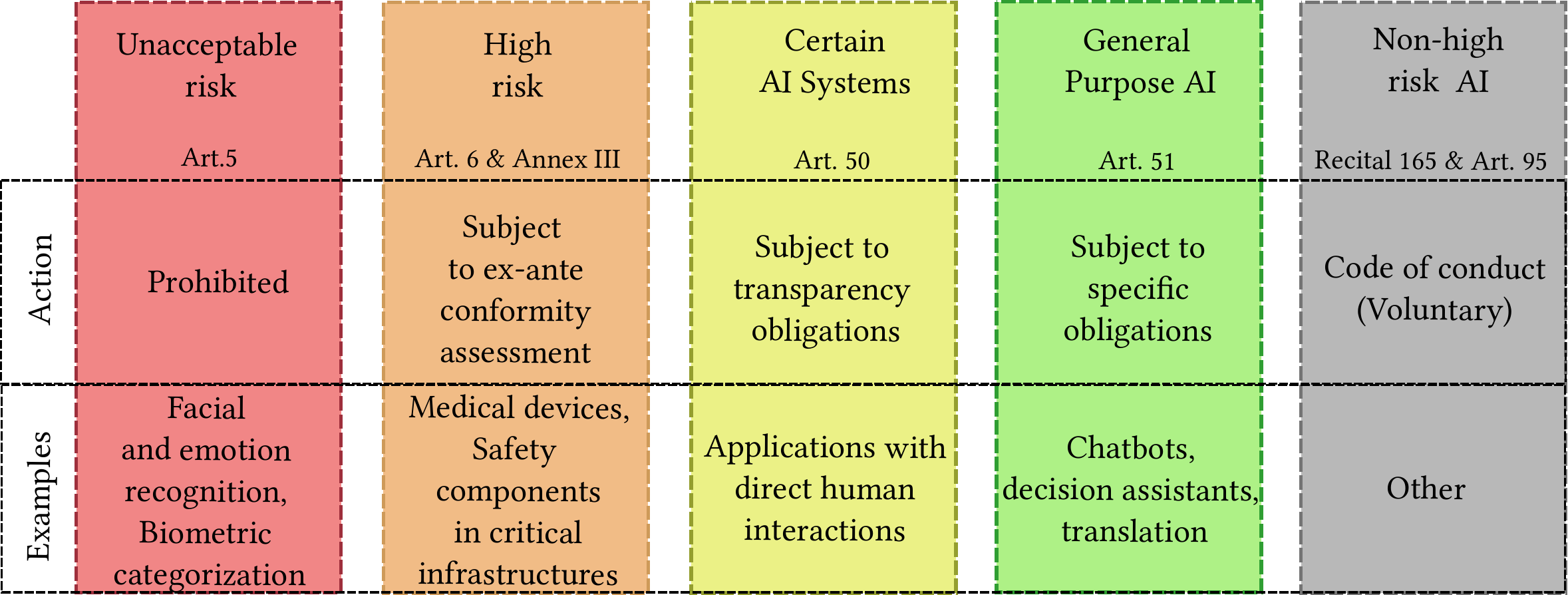}
    \caption{Risk categorization under the AI Act}
    \label{fig:regulatory_categories}
    \Description{The figure presents 5 types of risk according to AI Act. There are 5 bars: the first bar in red defines unacceptable risk level according to Art. 5 of AI Act. No actions are allowed for this risk type. Examples of this risk type are faction recognition systems and biometric categorization. The next orange bar presents High-risk systems, according to Art. 6. These systems are subject to ex-ante assessment. Examples of such systems are medical devices and safety components for critical infrastructures. The third bar is certain AI systems, according to Art. 59. These systems are subject to transparency obligations and are systems with direct human interaction. The fourth bar in green is general-purpose AI systems that have specific obligations. These systems are chatbots and decision assistants. The last bar in grey is non-high-risk AI systems.  }
\end{figure*}

It defines \ac{AI} in Art. 3 point (1) as \textit{a machine-based system designed to operate with varying levels of  autonomy, that may exhibit adaptiveness after deployment and that, for explicit or  implicit objectives, infers, from the input it receives, how to generate outputs such as 
predictions, content, recommendations, or decisions that can influence physical or virtual environments}.
The European Commission did not choose a sector-specific \ac{AI} regulation but rather an all-encompassing and principle-based Regulation, including possible categories of \ac{AI} application in all sectors: unacceptable risk, high risk, certain types of \ac{AI} with transparency obligations, general purpose \ac{AI}, and non-risk risk \ac{AI}, as presented in Figure \ref{fig:regulatory_categories}. The cases of \ac{AI} with an unacceptable level of risk are laid down in Art. 5 of the \ac{AI} Act, followed by the high-risk areas in Art. 6 and in Annex III. Certain types of \ac{AI} systems are subject to transparency obligations according to Art. 50, followed by the general purpose AI Models, subject to specific obligations, listed in Art. 51. The remaining categories of non-high risk AI are defined in the recital (165) and in Art. 95 and can voluntarily comply with the Code of Conduct. The different risk categories implied a different set of rules and obligations, which \ac{AI}-developers would have to follow.

\subsection{AI as Safety Components}

The energy sector has been categorized as a high-risk field of application for \ac{AI} technologies if \ac{AI} technology is being used for safety components in the management and operation of critical infrastructures, including the digital critical infrastructures. These are defined in Art. 2 point 4 and in Annex I of the Directive (EU) 2022/2557 \cite{eu-2022-2257} and in Annex III point 2 of the \ac{AI} Act. High-risk \ac{AI} systems in the energy sector are referred to in Annex III point 2 of the \ac{AI} Act as: \textit{(a) \ac{AI} systems intended to be used as safety components in the management and operation of road traffic and the supply of water, gas, heating and electricity}. Such safety components are further defined in point (55) of the \ac{AI} Act, \textit{Safety components of critical infrastructure, including critical digital infrastructure, \textbf{are systems used to directly protect the physical integrity of critical infrastructure or health and safety of persons and property but which are not necessary in order for the system to function}. The failure or malfunctioning of such components might directly lead to risks to the physical integrity of critical infrastructure and thus to risks to the health and safety of persons and property. Components intended to be used solely for cybersecurity purposes should not qualify as safety components. Examples of safety components of such critical infrastructure may include systems for monitoring water pressure or fire alarm controlling systems in cloud computing centers.} This definition resembles the safety component definition in the Art. 2(b) of the Machinery Directive, where a safety component was described as \textit{\textbf{not necessary for the system to function}} \cite{eu-2006-42}. Such a definition is too product-oriented and ignores the fact that there are elements in the digital and critical energy infrastructure where safety and functioning are interdependent in the system's algorithm. Certain safety functions in generation plants, power grids, or digital infrastructures can be indispensable for the system's functioning and safety, e.g., using \ac{AI} for voltage and frequency-control mechanisms, stability assessment systems, and load balancing.

The \ac{AI} Act states that only if \ac{AI} technology is used in safety components needed for the safety and physical integrity but which are not necessary for the system to function, high risk can be identified. Such a limitation is very narrow and does not reflect the constantly evolving nature of smart grid \ac{AI} applications. How to regulate cases where \ac{AI} technologies might be interconnected and used in safety components for the protection and functioning of the whole system are not yet dealt with in the Act and will have to be clarified in an additional sector-specific delegated act. Until then, this will remain a grey zone, leaving room for interpretation for \ac{AI} providers. A potential consequence could be that \ac{AI} applications developed for the safe operation of critical infrastructure, such as voltage or frequency control, but which do not meet the safety component definition under the Act, will simply not be covered by the \ac{AI} Act and will not be obliged to meet the High-Risk \ac{AI} compliance and accountability obligations. This is questionable in light of the potential risk implications for the system operators and users of the critical infrastructure \cite{Niet}.

\subsection{Problem Formulation}

As a result, the \ac{AI} Act does not provide sufficient regulation action for all the variety of use-cases of \ac{AI}-based services in the power grid. The definition of the safety component is capturing only a subset of possible functions that can be realized using \ac{AI}, while other potential \ac{AI}-based services are not covered. Since unreliable usage of \ac{AI} in critical infrastructure is a major public safety concern, additional actions are required to guarantee that the development and implementation of the \ac{AI} has been carried out responsively. In IEC 61508 \cite{iec61508-1}, qualitative and quantitative approaches have been proposed in the safety-relevant systems to assign risk levels to the safety functions. In \cite{diemert2023safety}, a similar approach was proposed for general purpose \ac{AI} functions. Along with qualitative parameters, such as non-determinism, a quantitative framework for risk assessment and definition of rigor activities should be provided. Developing and legislating such safety standards for \ac{AI} is a complex but necessary task to be completed in the next decades. However, already nowadays, actions are required to provide granular control of the developing \ac{AI}-based services. A delegated act could be introduced and propose a methodology to supervise \ac{AI}-based service from the planning till the commission phase. Such supervision can be quantified through a mechanism of service accountability. Accountability is usually defined through its core dimensions: explainability, audibility, and trustworthiness. However, the interpretation of accountability varies across different domains \cite{boch, kacian}. A sector-specific accountability for \ac{AI}-based service should be defined, described, and quantified to overcome it. In this work, an accountable \ac{AI}-based smart grid service is defined, and an approach to preserve overall accountability through accountability of the separate development and deployment processes is presented.

\section{Defining Accountability}

\label{sec: definitions}
 
This section discusses the existing definitions of accountability in different sources: in \ac{AI} Act, in general and smart grid-specific literature. An approach to provide a quantifiable definition of accountable \ac{AI}-based smart grid service is presented. 

\subsection{Limitations of Definitions in the AI Act}
The current \ac{AI} Act demands accountability from \ac{AI}  applications in different ways. Accountability was already identified by the Commissions Expert Group on \ac{AI} in 2019, as a requirement for trustworthy \ac{AI} \cite{AI-Ethics}. In the \ac{AI} Act itself, accountability is not defined but
can be found partially hidden in terms of transparency, explainability and interpretability. These terms can be found in different Articles of the Act and concern primarily \ac{AI} providers' obligations regarding high-risk applications. The following non-exhaustive list of examples shows the legislators' intention to ensure transparency of \ac{AI} High-Risk systems through an ex-ante conformity assessment of \ac{AI} applications.
One clear example is the record-keeping requirement. As stipulated in Art. 12 point (1) and Art. 19 of the Act, automatic recording of events should be possible, illustrating the need for traceability of \ac{AI}. Interpretability and transparency requirements are to be found, among others, in Art. 13 and in Section 3 of the Act. It states that High Risk \ac{AI} applications shall be designed so that their operation is sufficiently transparent to enable users to interpret and use the system's output. Further, the requirement for human oversight of \ac{AI} decisions and in \ac{AI} design in general are to be found in Art. 13 (3)(d) and Art. 14 of the Act. Lastly, data quality of \ac{AI} is to be ensured through a quality management system and documentation requirements in Art. 17 and 18. 

Thus, the Act does not define \ac{AI} accountability as a separate term but requires High-Risk \ac{AI} providers and deployers to follow different obligations, leading to more trustworthiness, irrespective of their application sector. The described approach covers the most crucial aspect of \ac{AI} development and deployment. However, it requires more granularity for recommended and undesired techniques at each step for specific domains, e.g., for the smart grid.

\begin{table*}[ht!]

\small{}
   \centering 

\begin{tabularx}{\textwidth}{p{2cm}>{\raggedright\arraybackslash}p{10cm}>{\raggedright\arraybackslash}p{5cm}}

    \textbf{Requirements} & \textbf{Characterization} & \textbf{ Resources} \\

\hline

 Responsibility & The obligation of a system or individual to follow procedures or standards
&  {\textbf{\cite{liu}}} {\textbf{\cite{volkova}}} \cite{nissen} \cite{boven} \cite{latifat}  \cite{liuha} \cite{lechter} {\textbf{\cite{bhatta}}} \cite{srini} \cite{loi} \cite{koppel} \cite{slota} \cite{oluwaseyi}  \cite{boch} \\

Explainability & Ensure understanding and justification of AI systems, processes and decisions
&  {\textbf{\cite{volkova}}} \cite{william} \cite{busuioc} \cite{liuha} \cite{smith} \cite{lima} \cite{pang} \cite{boch} \cite{srini} \\

 Transparency & Social-ethically centered information accessibility 
&  \cite{william} \cite{koppel} \cite{busuioc} \cite{tabassi} \cite{loi} \\

 Answerability & Design of a process or system that allows the clarification of an action
&  \cite{nissen} \cite{william} \cite{busuioc} \cite{noveli} \cite{loi} \\

 Trustworthy &  A property of acting ethically and demonstrating integrity and trust 
&  {\textbf{\cite{volkova}}} \cite{nissen} \cite{liuha} \cite{pang}\\

 Auditability & Verification if processes follow policies, standards or regulations
 & \cite{william} \cite{liuha} \cite{pang} \cite{srini}\\

 Moral \&  Ethical & Enhancing legal and constitutional requirements in system architecture
&  \cite{srini} \cite{cooper} \cite{oluwaseyi}\\

 Traceability &  The ability to track a process through several phases of a system production 
& {\textbf{\cite{liu}}} {\textbf{\cite{yang}}} \cite{srini} \\

 Punishment & Consider penalty for any harmful act of a system 
& {\textbf{\cite{liu}}} \cite{nissen} \\

 Property & Consider any attribute or characteristic as a feature for defining a system 
&  \cite{liuha} \cite{kacian}\\


 Interpretability & Ensure that a model and its outcomes are clear and simple to understand
&  \cite{william}\\

 Attributability & Identify the role or function of each component in the system
&  \cite{pang} \\

 Recordability & Validate that all actions are entered into the system and documented
&  {\textbf{\cite{liu}}}\\

 Controllability & To provide authority and accessibility measurement for a process or system
&  \cite{koppel}\\


\bottomrule
    \end{tabularx} 

    \caption{Accountability requirements in the literature, including smart grid domain-specific work (in bold)}
    \label{tab:my_label2}
\end{table*}

\subsection{Accountability in Literature}

The general literature on \ac{AI} lacks a precise definition of accountability for \ac{AI}-based systems due to its relation to several other aspects. Nissenbaum defines it as a requirement to provide information regarding a performed action, explain why it was taken, and undertake a follow-up action, which could include various responses such as punishment or penalty \cite{nissen,william}. 
In \cite{liuha}, the authors remark that ``\textit{In general, accountability for \ac{AI} indicates how much we can trust these \ac{AI} technologies and who or what we should blame if any parts of the \ac{AI} technologies perform below expectation. It is about a declaration of responsibility. It is not trivial to explicitly determine the accountability for \ac{AI}. On the one hand, most \ac{AI}-based systems act as black-box, due to the lack of explainability and transparency. On the other hand, real-world \ac{AI}-based systems are very complex and involve numerous key components, including input data, algorithm theory, implementation details, real-time human control. These factors further complicate the determination of accountability for \ac{AI}. Although difficult and complex, it is necessary to guarantee accountability for \ac{AI}. Auditability, which refers to a set of principled evaluations of the algorithm theories and implementation processes, is one of the most important methodologies in guaranteeing accountability.}'' In \cite{srini}, it is stated that ``\textit{Six main dimensions can be associated with accountability according
to the goals and needs of the different stakeholders: responsibility,
justification, audit, reporting, redress, and traceability, and in this regard,
there are several excellent works that focus on the aforementioned
specific dimensions.}'' In \cite{slota}, the research indicates that ``\textit{Accountability in \ac{AI}, thus, becomes less about how \ac{AI} is built, and more about how it is understood, legislated, and regulated.}''

Keywords derived from the definitions include ``responsibility'', ``transparency'', ``auditability'' and ``explainability''. These terms represent the prerequisites for accountability in \ac{AI}-based systems, as they contribute to developing and implementing systems that prioritize trustworthiness, fairness, and responsibility for all stakeholders. Table \ref{tab:my_label2} extends this discussion and represents a collection of resources regarding accountability definition and related dimensions. This collection comprises 24 scientific articles defining accountability or highlighting its various dimensions, e.g., ``responsibility'' is mentioned in 14 analyzed works, and ``explainability'' -- in 9.

In the energy domain, accountability has not yet been sufficiently covered. In \cite{liu}, the authors state that ``\textit{Accountability is required to further secure the smart grid in terms of privacy, integrity, and confidentiality. Even if a security issue presents itself, the built-in accountability mechanism will find out who is responsible for it. Once detected, some problems can be fixed automatically through the predefined program, while others may provide valuable information to experts for evaluation. In essence, accountability means that the system is recordable and traceable, thus making it liable to those communication principles for its actions}.'' In \cite{volkova}, it is argued that ``\textit{Accountability of the AI-based services should play a central role where explainability approaches are not applicable, e.g., when a decrease in performance cannot be avoided. Accountable AI also increases trustworthiness of the implemented services by different stakeholders and end users (\dots)}.''

\begin{table*}[htb!]
  \centering
\small{}

\begin{tabularx}{\textwidth}{p{2.5cm}>{\raggedright\arraybackslash}p{6cm}>{\raggedright\arraybackslash}p{8.5cm}}

\textbf{Step} & \textbf{Possible risks to accountability} & \textbf{Examples of preservation methods} \\
 \hline

Dataset Generation & Lack of planning; Unreliable data sources; Induced errors; Insufficient or redundant data; Irresponsible data provider & Clear requirements to  accuracy, completeness, and consistency; \newline Planned and prepared data collection process; SLA for data quality;  Repeated collection from the same system in the presence of bias  \\
Digital Twin & Data errors; Bias; Incorrect assumptions; Oversimplification in digital twin model; \newline Incomplete data; Not representative data & Record of assumptions in twin model; Ensured correctness of the twin system; Ensured interoperability for digital twin and real system \\
SLA \cite{ardagna2023big} & Lack of clear requirements for data assurance  & Regulatory implementation of SLA for all iterations of service development \\
Data Storage & Loss of data; Loss of access to data; \newline Lack of integrity (unreliable storage) & Clear data storage requirements; Regulation regarding data storing period between power grid and service provider; Enabled access to data for all required parties; Ensured security; SLA for data storage \\
Data Induction & Leakage of irrelevant and private data & Only necessary data collected; Regulated and recorded operations on data; Data aggregation on edge; Data obfuscation \\
\bottomrule

    \end{tabularx}

  \caption{Accountability issues in data collection}
  \label{Table: Data collection}
\end{table*}

\subsection{Deriving a Definition}
\label{sec: our-definition}

A comprehensive definition would support the development of an accountability framework and should encompass various aspects of accountability relevant to different phases of the service life. The other challenge is quantifying accountability and measuring the risks if the \ac{AI} does not fulfill accountability recommendations. Providing a cross-sector unified definition of accountability is only possible at a high level of abstraction.

Thus, the definition of accountability should be directly applicable to the target service or process. In this regard, an accountable \ac{AI}-based smart grid service can be defined as service that is: \textit{1) conceptualized and developed according to the sector-specific regulations and with clear identification of roles and responsibilities of all involved parties 2) regulated over the whole life cycle from planning to commissioning to minimize risks of an incorrect behavior in each development phase by tracing the impact of each step on the service outputs, recording all the decisions and actions performed during the development and deployment as well as responsible parties, guaranteeing accessibility of collected and processed data for involved parties 3) monitored during the operation time with granular insight into the impact of design choices on the output}. It is important to note that explainability \cite{machleva} has been explicitly not used in this definition since its human-centered nature and the general tendency for reduced performance due to model simplification are not applicable for some of the power grid services \cite{boch, volkova}. In the case of accountability, complexity is seen as the source of potential design and development imperfections and related risks but is acceptable.

\section{Accountability Preservation in AI-Based Service Development}
\label{sec: framework}

Essentially, \ac{AI}-based service accountability, defined in Section \ref{sec: our-definition}, should be preserved by a technical framework, which should cover all the life cycle phases. The general phases are planning, development, and operation. This section focuses on the development phase, which involves data collection and storage, data preprocessing, feature selection and extraction, and utilizing learning algorithms \cite{steidl2023pipeline}. A wide range of possible techniques characterizes each of these steps. This section discusses exemplary major steps and related accountability risks and omits some smaller ones, e.g., in data engineering  \cite{nazabal2020data}. Strategies to preserve accountability in the planning, operation, and commissioning phases and business-related aspects will be discussed in future work. The early insight into accountability preservation during the operation phase is discussed in \cite{volkova}.

\subsection{Data Collection and Correction Phases}

The established and accountable data collection and correction process is the initial step of accountability assurance. In this regard, the smart grid is a special case since the data can have different origins: assembled from the measurements of the real system, from a digital twin, or from software simulation, each with its specific accountability issues. General accountability issues of this step are listed in Table \ref{Table: Data collection}.

In the data collection step, data source discovery should be associated with a detailed understanding of the behavior of the system components. Data collection for smart grid applications should consider the physics of the processes, e.g., for voltage and frequency control applications. Furthermore, sensor data collection can be associated with many induced errors due to sensor aging and losses during data transmission. The associated risks should be well documented at this stage and forwarded to the data correction methods. Preliminary data sampling can help to identify and fix faulty components. Sampling data from the digital twin \cite{SLEITI20223704} is a concern for accountability due to the complexity of the twin system modeling and introduces the digital twin provider as a responsible party.  

For many grid services, the data is heterogeneous with varying resolutions, mostly asynchronous, and is stored in different formats (raw or processed) at different locations \cite{bhattarai}. Collected and processed data should be stored throughout service deployment and additional time after commissioning to preserve accountability. Access to the original and training data and clear documentation of the data origin, collection process, and format can enable post-factum analysis of incidents and study the impact of data on the service output. Careful and planned data storage is also required to guarantee an accountable re-training process, especially when data is shared between different stakeholders in the energy sector. A well-defined and extensive \ac{SLA} between the parties can be used to introduce responsibility in data utilization. A regulatory framework may support \acp{SLA} formulation for explicit usage in smart grids and performance metrics for data quality.

\begin{table*}[t]
  \centering

\small{}
\begin{tabularx}{\textwidth}{p{3cm}>{\raggedright\arraybackslash}p{6cm}>{\raggedright\arraybackslash}p{8cm}}

\textbf{Step} & \textbf{Possible risks to accountability} & \textbf{Examples of preservation methods} \\
\hline

 Data  Cleaning \cite{zainab2021datacleaning} & Incorrect and/or undocumented pattern detection; Loss of data; Loss of granularity & Planned and recorded actions; Validation procedures; \newline Quality assurance metrics \\
 Filtering  & Unregulated rules; Loss of data  & Planned and recorded filtering criteria; Validation procedures; \newline  Quality assurance metrics  \\
Uncertainty  Mitigation  & Loss of data;  Loss of correlations in data &  Clear definition of uncertainty types; Identification of the tolerable uncertainty levels;  Validation procedures \\
 Anomaly Detection  & Loss of data;  Loss of correlations in data  & Clear definition of anomalies; Evaluation of the anomaly scope; \newline  Validation procedures; Quality assurance metrics  \\

 Dimensionality Reduction  & Loss of data due to inaccurate reduction; Loss of correlations in data; Bias; Overfitting;   \newline Complexity of composite dataset & Planned and recorded actions; Responsible method selection  \\

 Synthetic Data \cite{razghandi2023smart} & Infeasible data; Leakage of data origin; Incorrect \newline data syntheses; Loss of complex dependencies;  & Justification for synthetic data selection; \newline  Validation of data harmonization  \\

 Augmentation \cite{zhang2020data} & Uncontrollable infusion of additional data;  \newline Loss of data &  Recorded augmentation strategy; \newline  Validation of data consistency \\

Labeling \cite{paleyes2022challenges, ashktorab2021ai} & Inconsistent labeling rules; Error and loss of accountability through AI-tools & Avoidance of non-deterministic tools; Recorded strategy for labeling; Validation within each label \\

Data Anonymization  &  Insufficiently described actions; Data loss; \newline Loss of relations between the data points & Avoidance of sensitive data during the collection phase; \newline  Anonymization during the collection; Recorded rules and assumptions \\

\bottomrule

    \end{tabularx}%

  \caption{Accountability issues in data quality}
  \label{Table: Data Quality}
\end{table*}

\subsection{Data Quality Assurance and Preprocessing}

Poor data quality in the training process can directly affect service performance, and the original reason for such behavior will be hard to identify. Data preprocessing methods are mainly employed to address issues inherited from the data collection process (uncertainties \cite{hariri2019uncertainty}, errors, lack of data). These methods may be one of the core sources of accountability risks, as they involve techniques that transform the original data. The risks related to removing, adding certain data, or aligning data points impact the developed model and should be carefully identified. By data dispersion, several data sources with different schema or conventions are merged into a single dataset. The differences in data structures and types and the various tools required to enable data integration need special attention. The development methodology should be capable of finding a trade-off between the necessary preprocessing steps and the risk to accountability these introduce due to their complexity and required assumptions. Data dispersion from multiple sources is undesired for accountability preservation in smart grid services, and a unified data collection plan should be introduced. An overview of some of the potential technical processes can be found in Table \ref{Table: Data Quality}.

Smart grid data may have hard-interpretable labels depending on the type of service. Each method of data labeling has an impact on accountability. Manual labeling is prone to human error, while automated labeling tools usually include different \ac{AI} methods. Such labels can inherit all issues from the accountability of the tools \cite{ashktorab2021ai}. Accountability for this step can be preserved by clearly planning the labeling process and continuous validation and refining through feedback loops. Data leakage and dispersion can threaten the system's accountability during the filtering and labeling processes. By data leakage, target variables (irrelevant or personal data) are leaked in the training. When leakage happens, the model retains characteristics connected with irrelevant data without explicitly including those features in the model. The accountability of this process should be preserved by careful planning, early issue detection, and documentation of the completed steps and used tools \cite{paleyes2022challenges}. Since the fusion of private and system data is common for the power grid, numerous techniques can be used to anonymize data. Anonymization reduces the granularity and accuracy of the data, leading to the loss of the complex correlations between data points. This can lead to accountability issues since the model's output is unreliable due to the lack of critical data or data patterns. Time-series data in the power grid should be handled delicately to preserve diffused behaviors over larger time scales.

Data augmentation methods, including synthetic data, resolve the issues related to limited data volume or unbalanced data. However, synthetic data is vulnerable to information leakage from original data and inaccuracies. Moreover, capturing outliers and low-probability events in synthetic data is challenging \cite{paleyes2022challenges}. A synthetic data generator may not replicate statistics accurately. The same applies to the common for smart grid digital twin data application in real systems \cite{jafari2023review}. Thus, low-probability events may not be evident or have different patterns in the twin and real system behavior.

\begin{table*}[th!]
  \centering

\small{}

\begin{tabularx}{\textwidth}{p{3cm}>{\raggedright\arraybackslash}p{6cm}>{\raggedright\arraybackslash}p{8cm}}

\textbf{Step} & \textbf{Possible risks to accountability} & \textbf{Examples of preservation methods} \\

\hline

Feature Selection \cite{dhal2022comprehensivefeaturessel} &  Improper usage of selection techniques; Usage of too complex techniques & Preference towards deterministic methods  \\

Feature Transformation & Improper scaling and encoding  & Preference towards simple features; Recorded and justified rules for complex features \\
Model Selection & Insufficient analysis of model applicability; Insufficient preliminary analysis of model benefits & Model selection adequate problem requirements; Risk assessment for each candidate model \\
Model Training & Insufficient cross-validation; Inaccurate processing of hyperparameters; Model bias & Validation metrics for each training cycle; Clear track of model performance evolution  \\
Hyperparameter \newline Tuning \cite{probst2019tunability} & Insufficient parameters; Unnecessary probabilistic models to find hyperparameters  & Search space analysis; Guaranteed reproducibility of the selection process  \\

Model Validation & Not representative training set; Insufficient metrics; Lack of analysis  &  Recorded configuration of validation experiments; Analysis of data preprocessing decisions in validation phase \\
\bottomrule
    \end{tabularx}%

  \caption{Accountability issues in model training}
  \label{Table: Model Training}

\end{table*}

\subsection{Model Training and Validation}

After the data collection and preparation phase, the datasets are ready for training. From this point, the algorithm will inherit all the remaining issues in the data. Accountability of the data preparation phase allows access to the used assumptions and methods, especially if the dataset is being used by a different party or reused to develop multiple services. Table \ref{Table: Model Training} presents primary training steps and related risks. 

The model selection and training are based on the objective and available features. The feature engineering process usually includes extraction, selection, and transformation phases. Accountability risks arise from the feature selection process due to incorrect formulation of the objectives and assumptions, resulting in feature extraction and transformation flaws. Improper feature selection techniques can lead to losing important features while eliminating unsuitable ones. It can fail to achieve dimensionality reduction and overlook significant feature correlations. The feature transformation methods may require dividing a feature or merging two or more features to build a new one using various methods. Applying the transformation methods changes the essence of data and may introduce biases. Algorithmic bias appears when mathematical rules highlight specific attributes over others in relation to a target variable  \cite{Bantilan2018bias}. The feature engineering process should be carefully planned and documented to track the steps of the process for the ex-post analysis. Feature selection and labeling also allow methods to approximate the model locally with other interpretable models and understand the influence of features on service output. 

 Accountability in the model learning phase should be preserved for three main steps: model selection, training, and hyperparameter selection. Justification of the model selection should be provided concerning the complexity of the problem and the application scenario. The increased complexity of the selected algorithm impacts accountability as it deals with sensitive hyperparameters. Tuning requires a knowledge of the search space, which is not always achievable due to complexity. Further constraints from the deployment environment may also introduce bounds on the hyperparameter selection \cite{paleyes2022challenges}. Accountability issues arise when the hyperparameter's choices and bounds are insufficiently justified. Inaccurately tuned hyperparameters result in overfitting, underfitting, or bias. 
 
 The training process is initiated after carefully implementing and selecting the hyperparameters, usually in parallel with the performance evaluation. Accountability of this step is defined through extensive validation with a feasible selection of the validation metrics that should cover accuracy, bias, and fairness. Accountability may be compromised if the chosen metrics fail to consider bias issues or detect parameter errors for the training and unseen test sets. However, the test set can further compromise accountability by not being representative or infected with training data. Such risks will be explicitly evaluated in a future use-case study.

\section{Conclusion}
Additional measures are needed to ensure responsible development and deployment for reliable \ac{AI} usage in critical infrastructures. This paper analyzes the current \ac{AI} Act and demonstrates its limited granularity in considering technical aspects of the smart grid, e.g., in safety component definition. A delegated act could include a methodology for supervising critical AI-based services in the energy sector, closing the current gap in the AI Act. An accountability-based framework is suggested as a fundamental part of a delegated act. A literature survey of existing definitions of accountability identifies the most significant accountability dimensions, and sector-specific quantifiable accountability definition for \ac{AI}-based smart grid service is derived. The quantification of the accountability risks can be narrowed down to each phase of the \ac{AI}-based service development from planning to commissioning. This work discusses accountability risks for data collection, preprocessing, and model training phases. The future work will consider implementing the discussed accountability preservation in an exemplary \ac{AI}-based service, underlining the impact of design decisions on operation and the assigning responsibility of different parties.

\begin{acks}
This work was funded by project ``EnerSat (Satellitengest\"utzte Kommunikation f\"ur das Energiesystem)'' that has received funding from BMWK (Bundesministerium für Wirtschaft und Klimaschutz) under funding indicator 50RU2303C and by the European Union’s Horizon 2020 research and innovation programme under grant agreement No. 957845. The authors would like to thank Dr. Brenda Espinosa Apr\'aez from the Tilburg University, the Netherlands, for insightful discussions on the AI Act.

\end{acks}

\bibliographystyle{ACM-Reference-Format}
\bibliography{sample-base}

\end{document}